\documentclass[10pt,twocolumn,letterpaper]{article}

\usepackage{cvpr}
\usepackage{times}
\usepackage{epsfig}
\usepackage{graphicx}
\usepackage{amsmath}
\usepackage{amssymb}
\usepackage{enumitem} 
\usepackage{refstyle} 
\usepackage[square, sort, numbers]{natbib} 

\usepackage{xcolor}
\usepackage{xcolor,colortbl}
\usepackage{subcaption}
\usepackage{bm}

\usepackage[pagebackref=true,breaklinks=true,colorlinks,bookmarks=false]{hyperref}

\usepackage{algorithm,algcompatible,amsmath}
\algnewcommand\INPUT{\item[\textbf{Input:}]}%
\algnewcommand\OUTPUT{\item[\textbf{Output:}]}%

\newcommand{\nwidth}{$N$-Width }
\newcommand{\Ebb}{\mathbb{E} }
\newcommand{\braces}[1]{\{#1\}} 
\newcommand{\vect}[1]{\bm{#1}}
\newcommand{\mat}[1]{\bm{#1}}

\cvprfinalcopy 

\def\cvprPaperID{29} 

\ifcvprfinal\fi
\begin{document}

\title{Any-Width Networks}

\author{
{Thanh Vu 
    \qquad Marc Eder
    \qquad True Price
    \qquad Jan-Michael Frahm}\\
University of North Carolina at Chapel Hill\\
Chapel Hill, NC\\
{\tt\small \{tvu, meder, jtprice, jmf\}@cs.unc.edu}
}

\maketitle
\thispagestyle{empty}  

\begin{abstract}
   Despite remarkable improvements in speed and accuracy, convolutional neural networks (CNNs) still typically operate as monolithic entities at inference time. This poses a challenge for resource-constrained practical applications, where both computational budgets and performance needs can vary with the situation. To address these constraints, we propose the Any-Width Network (AWN), an adjustable-width CNN architecture and associated training routine that allow for fine-grained control over speed and accuracy during inference. Our key innovation is the use of lower-triangular weight matrices which explicitly address width-varying batch statistics while being naturally suited for multi-width operations. We also show that this design facilitates an efficient training routine based on random width sampling. 
   We empirically demonstrate that our proposed AWNs compare favorably to existing methods while providing maximally granular control during inference.
\end{abstract}

\section{Introduction}

Recent advancements in convolutional neural networks (CNNs) have significantly improved both the speed and accuracy of state-of-the-art computer vision algorithms, in turn enabling the development of low-latency applications such as those for mobile platforms and autonomous vehicles. However, fast execution alone is not always sufficient. Time constraints can be situationally dependent for many applications, and there is an observable trade-off between speed and accuracy for many vision systems \citep{speed_acc_obj_detect}. For example, those in autonomous vehicles can have a limited, yet dynamic, computational budget that varies over time due to changes in the vehicle's velocity, competition for resources, and other factors. Thus, vision algorithms must not simply be fast, but also adaptive and flexible to varying budgets. This need is at odds with the typical architectural design of CNNs as monolithic entities. Yet, ideally, a resource-constrained CNN should allow the flexibility to vary its operation along a speed-accuracy curve at inference time.

\begin{figure}[t]
    \centering
    \begin{subfigure}[b]{0.45\linewidth}
        \centering
        \captionsetup{justification=centering}
        \caption*{\textbf{Proposed AWN \\*
        Architecture}}
        \includegraphics[height=5.5cm]{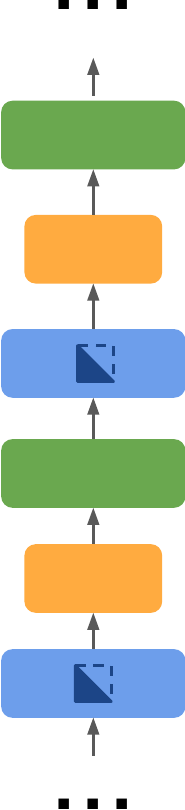}
    \end{subfigure}
    ~
    \begin{subfigure}[b]{0.45\linewidth}
        \centering
        \captionsetup{justification=centering}
        \caption*{\textbf{Slimmable \\*
        Architecture \citep{slimmable, slimmable_v2}}}
        \includegraphics[height=5.5cm]{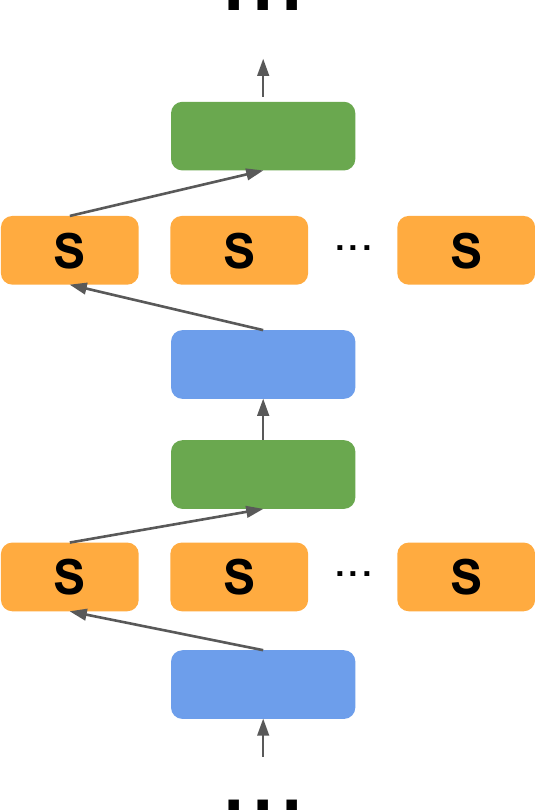}
    \end{subfigure}\\
    
    \vspace{1em}
    \begin{subfigure}[b]{\linewidth}
        \centering
        \includegraphics[height=0.6cm]{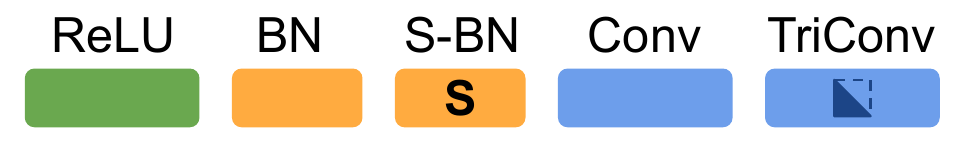}
    \end{subfigure}
    \caption{Our AWN architecture provides a single model for all widths of operation. We propose to use a lower-triangular convolution layer (TriConv), which permits networks to use traditional batch normalization layers (BN). In contrast, prior works \citep{slimmable, slimmable_v2} require a custom, switchable batch normalization layer (S-BN) to store separate versions of activation statistics for every desired width of operation.}
    \label{fig:arch-compare}
    \vspace{-1em}
\end{figure}

Many recent works have focused on improving efficiency by adjusting the number of layers, or \textit{depth}, of the network \citep{fractalnet, msdnet, adaptive-inference, adaptive-inference-cnn, blockdrop, decision_gates_cvprw18}. Fewer techniques \citep{slimmable, slimmable_v2, nestednet} have been proposed to modify the number of active channels at each layer, or \textit{width}, of a network to control the trade-off between speed and accuracy. Width manipulation not only reduces necessary computation, but it can also reduce a network's memory footprint 
as the number of intermediate feature maps is reduced when fewer channels are active. 
Recent width-modulation strategies have followed an \textit{\nwidth} design, training a single network to operate at a fixed set of $N$ modes that can be switched between at inference time. To function in this way, these networks typically store multiple copies of some elements of the trained model. For example, Slimmable Networks \citep{slimmable, slimmable_v2} maintain $N$ separate batch normalization layers per convolutional layer for use during inference, while Nested Nets \citep{nestednet} store multiple sparsity masks for each layer to mask out neurons according to the width of operation.

There are a few notable drawbacks to the \nwidth design, however. First, the layer widths are fixed after training, which limits inference-time control to a fixed set of predetermined ``switchable'' modes. If $N$ is small, the resulting model has only coarse control during inference, but if $N$ is increased to provide more fine-grained control, the memory footprint can increase significantly due to the need to store $N$ versions of certain layers of the model. \citet{slimmable} note
that each additional set of batch normalization parameters in their Slimmable Network can add roughly $1\%$ to the model size. While this is may not seem burdensome with a small number of modes, it quickly becomes prohibitive as $N$ grows. For example, training the light-weight MobileNetV2 \citep{mobilenetv2} at 20 widths would result in an $18.5\%$ increase in the model size. Additionally, at large values of $N$, training becomes inefficient and impractical due to the simultaneous training of $N$ sub-networks, which increases training complexity and time. \citet{slimmable_v2} address this concern by randomly sampling widths during training, which allows for simultaneous optimization of a large number of widths, but they still require a post-training step to accumulate batch statistics. Finally, the selection of functional widths must happen during training, before the network's speed-accuracy trade-off curve can actually be computed. Thus, we are forced into a causality dilemma wherein we must define the desired control over the widths of operation without  knowing how a network will respond at each width.

In this work, we rethink the \nwidth design in order to address these shortcomings. We propose \textit{Any-Width Networks} (AWNs), which provide maximally granular control over the width of operation and can be trained using an efficient algorithm based on random width sampling. AWNs leverage lower-triangular weight matrices to operate at any width without the need for switchable modules. This design explicitly removes the width-dependent variation in batch normalization statistics observed by \cite{slimmable} while still allowing AWNs to use a single batch normalization layer per convolution for all widths. Additionally, AWNs can be efficiently trained using a random width sampling strategy similar to \citet{slimmable_v2}, but without any additional post-training processing steps. Lastly, AWNs provide a smoother, more consistent trade-off curve at all widths than prior work, which lessens the need to know the exact trade-off curve during training.

We summarize our contributions as follows:
\begin{itemize}[topsep=0pt,itemsep=-1ex,partopsep=1ex,parsep=1ex]
    \item We present a theoretical and empirical analysis of the problem of varying activation statistics in multi-width networks.
    \item We introduce triangular convolutional layers, which not only explicitly address the varying statistics problem but are also naturally suitable for multi-width setup and enable any-width control.
    \item We propose the Any-Width Network (AWN) architecture, combining triangular convolution, standard batch normalization, and an efficient training algorithm to provide fine-grained control over speed and accuracy at inference-time.
\end{itemize}
\section{Related Work}
A number of prior works have examined the trade-off between model size and accuracy in network design. \citet{wide_resnet} study the performance of the popular ResNet architecture \citep{resnet} when varying the width of the network. \citet{speed_acc_obj_detect} investigate the speed-accuracy trade-off of various object detection algorithms to provide a guide for application-based model selection. \citet{mobilenetv1} and \citet{mobilenetv2} propose a family of fast, compact networks called MobileNets that provide levers such as width and resolution multipliers to appropriately scale the network for different applications. Other researchers have proposed various network pruning techniques to accelerate inference and reduce redundancy in deep CNNs. This line of work, also known as network compression, includes weight pruning \citep{learning-connections, deep-compression, training-sparse-net, compress-aware, learning-compress, nisp, dynamic-net-surgery}, channel pruning \citep{channel-pruning, runtime-neural-pruning, net-slimming, prune-filter, thinet}, layer skipping \citep{adaptive-inference, stochastic-depth, adaptive-inference-cnn, blockdrop}, and decomposition \citep{coreset-compress, group-approx}. While these methods are effective, the performance trade-off decisions therein are made through the network design or at training time. We are looking to enable this type of control at inference time instead.

Numerous previous approaches have considered modifying a network's depth as a mechanism for inference time control. One such design is to use early-exits or early-stopping where predictions can be extracted from early layers in order to meet a resource budget \citep{fractalnet, msdnet, anytime_adapt_loss, adaptive-inference}. Others dynamically adapt networks by skipping or dropping layers on the fly based on the network input \citep{adaptive-inference, adaptive-inference-cnn, blockdrop}. These efforts share the same overarching goal of our approach, but depth modulation is an orthogonal approach to resource-constrained control.

Most closely related to our efforts are methods that provide adjustable width operation during inference. \citet{nestednet} and \citet{slimmable} both propose to consider a single network as a collection of sub-networks in order to train a network to operate at different widths. These papers form the basis for the \nwidth design paradigm. \citet{nestednet} propose a connection pruning method to iteratively train a network for a predefined number of sparsity ratios to obtain a fixed set of inference paths or internal networks. They then jointly optimize them to learn a final ``$N$-in-1'' nested network. Conversely, \citet{slimmable} train a network for a fixed $N$ number of predefined widths and raise the issue of inconsistent activation statistics for multi-width networks. They propose a ``switchable'' batch normalization module to address the problem. \citet{slimmable_v2} build on this idea, proposing the Universally-Slimmable Network, which uses random width-sampling 
to efficiently train arbitrary widths. They also propose a post-training routine to accumulate activation statistics for each desired width of operation. While this approach provides viable inference-time control, it still relies on defining $N$ switchable modes for inference. In our work, we aim to provide consistent performance with maximally granular control. That is, we aim to provide assured performance at any width of operation.
\section{Varying Activation Statistics}\label{sec:varyingactivationstats}
Batch normalization \citep{batch_normalization} is a prolific and important technique that helps to stabilize and expedite network training and provides better network generalization. Due to these welcomed improvements, the operation is prevalent in many common network architectures \cite{resnet, googlenet, mobilenetv1, mobilenetv2}. The operation normalizes each activation within a layer to have zero mean and unit variance, according to running means and variances accumulated over all mini-batches during training. It also learns a pair of scale and shift parameters which it uses to tune the degree of normalization. At inference time, all parameters are fixed and used to approximate the true statistics. Key to this process is the assumption that we can obtain a reasonable model of the distributions of activation states during training. Yet this premise may no longer hold in a multi-width setting because the number of active features at each layer changes with the width.
%

\subsection{Why statistics vary between widths}\label{sec:why-stats-vary}
Consider a fully-connected network comprised of layers whose outputs are feature vectors with $m$ dimensions. Each layer has an associate weight matrix, $\vect{W}$. The $i$-th layer is computed via matrix multiplication as:
\begin{equation}
    \vect{y}_i^{\braces{m_i}} = \vect{W}_i^{\braces{m_i \times m_{i-1}}} \vect{x}_i^{\braces{m_{i-1}}}
\end{equation}
where $\vect{x}_i$ and $\vect{y}_i$ are the input and output, respectively, and the superscripts indicate the dimensionality of each vector or matrix. Note that $\vect{x}_i$ is typically the output of the previous layer, and hence has dimension $m_{i-1}$. For simplicity, we leave out bias and the non-linear activation function without loss of generality.

When the network is trained in a multi-width setting, the number of active features in a given layer, $k$, is changed as:
\begin{equation}
    k_i = \alpha m_i
\end{equation}
where $\alpha$ is a scalar \textit{width-factor} of the whole network and $m_i$ is the total number of feature dimensions, active and inactive, in that layer.
For example, in a multi-width network, a fully-connected layer with $k = 1$ is computed as:
\begin{equation}
\begin{split}
    \vect{y}^{\braces{1}} &= \vect{W}^{\braces{1}} \vect{x}^{\braces{1}}\\
    \begin{bmatrix} 
        y_{1} 
    \end{bmatrix}
    &= \begin{bmatrix} 
        w_{11} 
    \end{bmatrix}
    \begin{bmatrix} 
        x_1 
    \end{bmatrix}\\
    &= 
    \begin{bmatrix} 
        w_{11} x_1
    \end{bmatrix}
\end{split}
\end{equation}
and with $k = 2$, it is:
\begin{equation}
\begin{split}
    \vect{y}^{\braces{2}} &= \vect{W}^{\braces{2}} \vect{x}^{\braces{2}}\\
    \begin{bmatrix} 
        y_{1} \\
        y_{2}
    \end{bmatrix}
    &=
    \begin{bmatrix}
        w_{11} & w_{12} \\
        w_{21} & w_{22} 
    \end{bmatrix}
    \begin{bmatrix} 
        x_{1}\\
        x_{2}
    \end{bmatrix}\\
    &= 
    \begin{bmatrix} 
        w_{11} x_1 + w_{12} x_2 \\
        w_{21} x_1 + w_{22} x_2 
    \end{bmatrix}
\end{split}
\end{equation}

Architectures that use batch normalization will subsequently normalize $\vect{y}$ according to accumulated statistics before passing it to the next layer. However, traditional batch normalization assumes that a single distribution can model all states of a given feature. While this is understood to be an effective assumption for standard networks operating at a single-width, it is clear from this example that the distributions of a feature will not necessarily be consistent across different widths in a multi-width network. Consider a single feature, $y_1$:
\begin{equation}
y_1 = 
\begin{cases}
    w_{11} x_1  & \text{if } k = 1\\
    w_{11} x_1 + w_{12} x_2 & \text{if } k = 2 \\
    \ldots & \ldots
\end{cases}
\end{equation}
This feature's expected value will differ between widths:
\begin{equation}\label{eq:expval}
\begin{split}
    \Ebb[y_1^{\braces{1}}] &= \Ebb[w_{11} x_1] = \mu_1^{\braces{1}} \\
    \Ebb[y_1^{\braces{2}}] &= \Ebb[w_{11} x_1 + w_{12} x_2]\\
    &= \Ebb[w_{11} x_1] + \Ebb[w_{12} x_2]\\
    &= \mu_1^{\braces{1}} + \Ebb[w_{12} x_2]
\end{split}
\end{equation}
Clearly, both $y_{1}^{\braces{1}}$ and $y_{1}^{\braces{2}}$ should not be normalized according to the same distribution. Unless we can ensure that $\Ebb[w_{12} x_2] = 0$, there is no guarantee that traditional batch normalization \citep{batch_normalization} can properly model the activation statistics of a feature in a multi-width network.
%
%

\begin{table*}[t] 

\begin{center}
  \begin{tabular}{r|cccc|cccc}
  \hline
    &\multicolumn{4}{c|}{\textbf{Training Accuracy}} &\multicolumn{4}{c}{\textbf{Validation Accuracy}} \\
             & \textit{w=1.00} & \textit{w=0.75} & \textit{w=0.50} & \textit{w=0.25} & \textit{w=1.00} & \textit{w=0.75} & \textit{w=0.50} & \textit{w=0.25}\\
  \hline
  \hline
    Standard BN & 99.29\% & 99.21\% & 99.02\% & 98.59\% & 98.49\% & 98.78\% & 96.59\% & 98.49\% \\
    Switchable BN \citep{slimmable, slimmable_v2} & 99.30\% & 99.24\% & 99.03\% & 98.57\% & 99.05\% & 98.93\% & 98.91\% & 98.69\% \\
    AWN (ours)
        & 99.23\% & 99.19\% & 99.04\% & 98.50\% & 98.97\% & 98.87\% & 98.82\% & 98.43\% \\
  \hline
  \end{tabular}
\end{center}
\caption{Comparison of different methods to address width-varying statistics using the LeNet-3C1L network for MNIST digit classification. Even though the training accuracies all seem quite similar, we see that, during validation, the use of a standard batch normalization layer results in inconsistent performance. Note that our AWN with triangular convolution performs only slightly worse than switchable batch normalization \cite{slimmable, slimmable_v2} on this simple problem, but with a smaller memory footprint.}
\label{table:sharedbn_acc}
\end{table*}


\begin{figure*}[t]
    \centering
    \begin{minipage}{.94\linewidth}
        \centering
        \begin{subfigure}[b]{0.32\textwidth}
            \includegraphics[width=\textwidth]{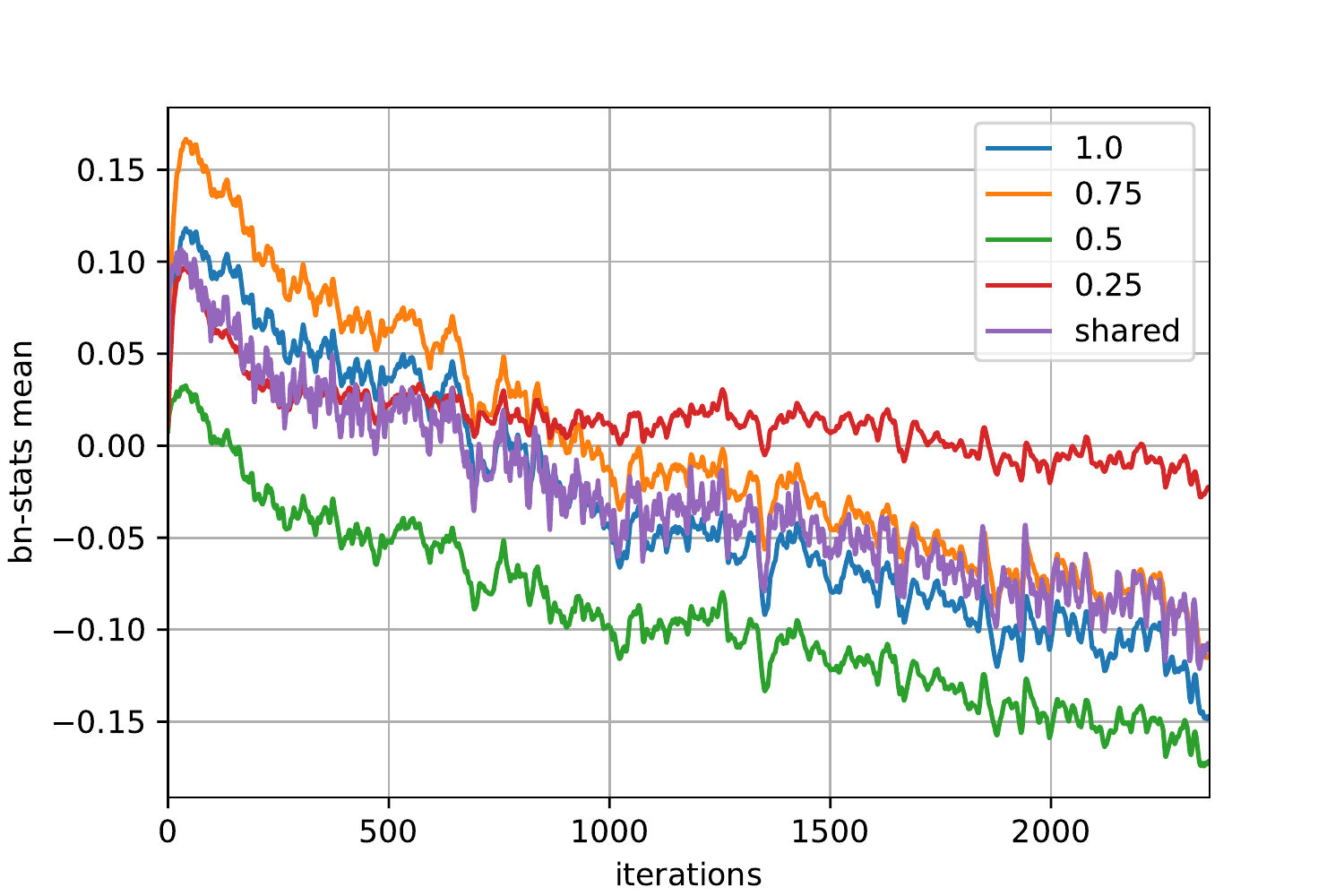}
        \end{subfigure}
        ~
        \centering
        \begin{subfigure}[b]{0.32\textwidth}
            \includegraphics[width=\textwidth]{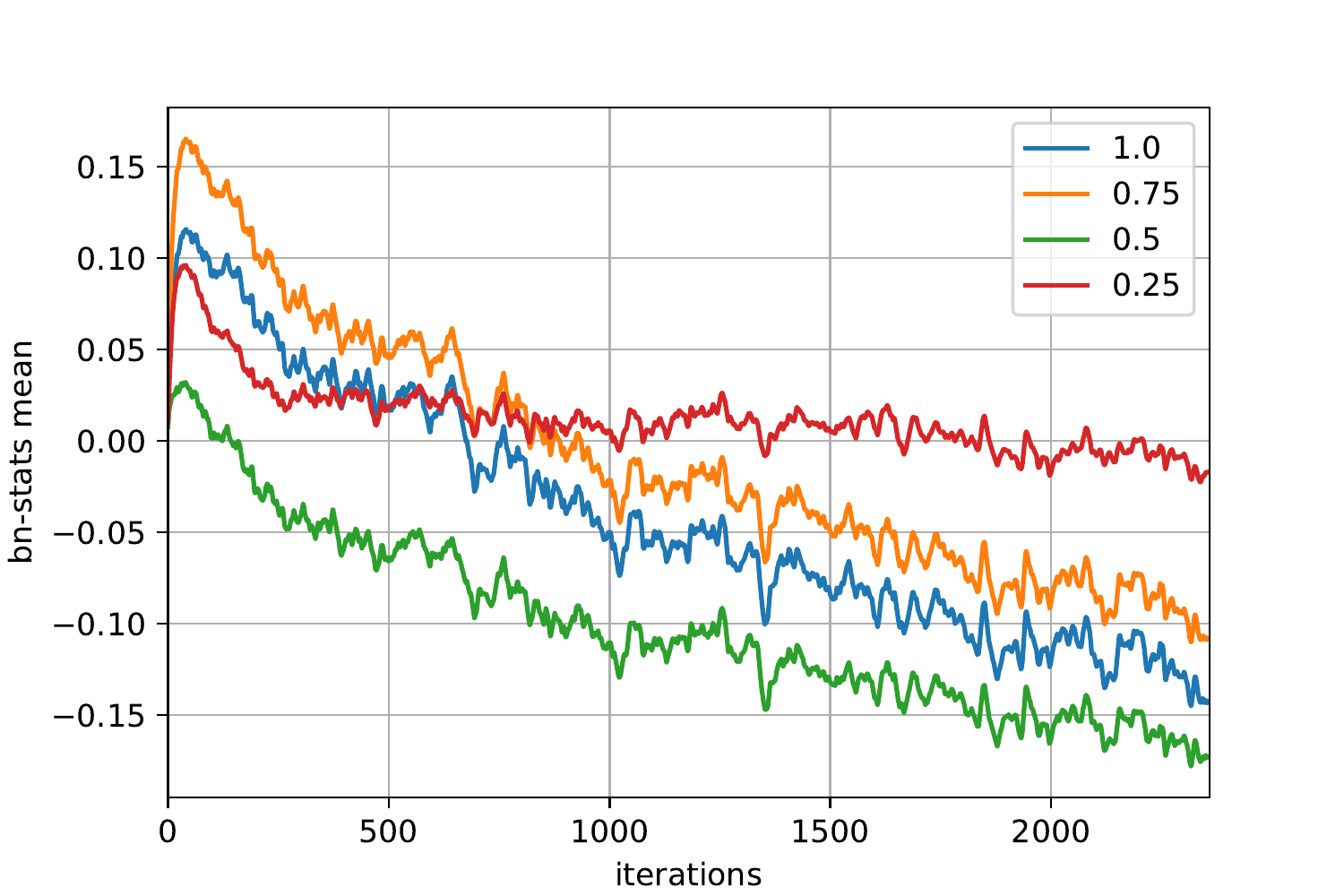}
        \end{subfigure}
        ~
        \centering
        \begin{subfigure}[b]{0.32\textwidth}
            \includegraphics[width=\textwidth]{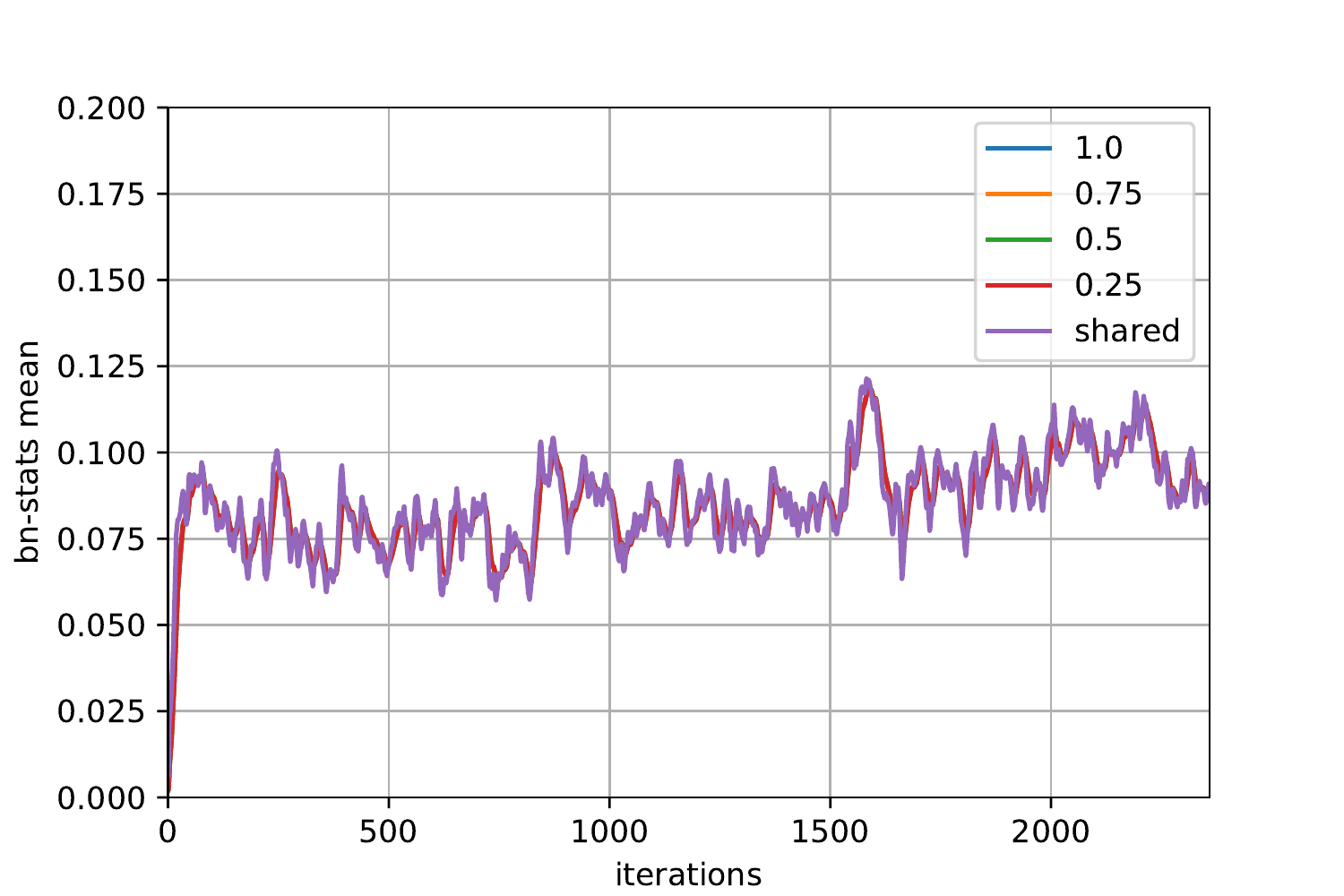}
        \end{subfigure}\\
    \vspace{4pt}
        \centering
        \begin{subfigure}[b]{0.32\textwidth}
            \includegraphics[width=\textwidth]{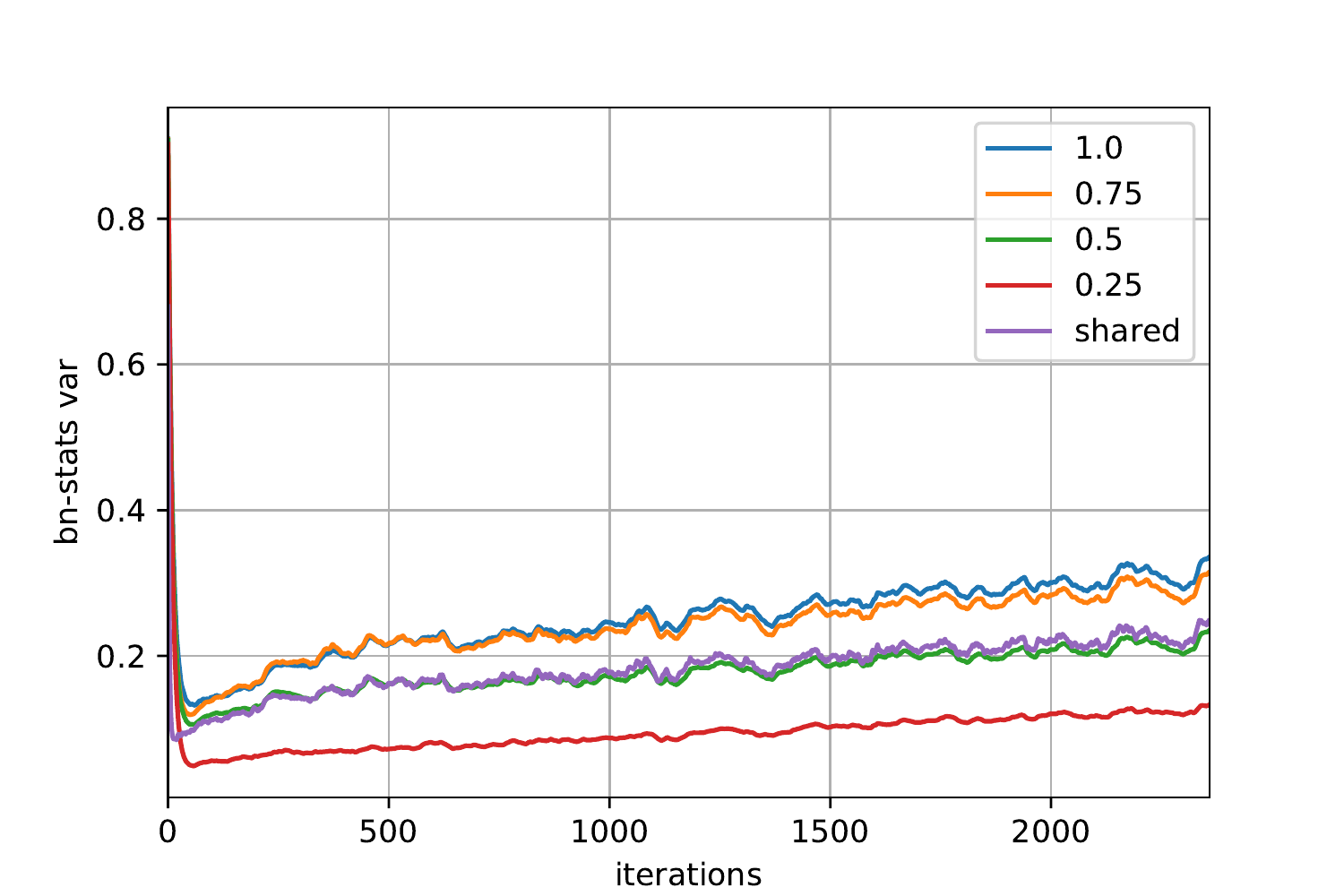}
            \caption{Standard BN}
            \label{fig:sharedbn_stats}
        \end{subfigure}
        ~
        \centering
        \begin{subfigure}[b]{0.32\textwidth}
            \includegraphics[width=\textwidth]{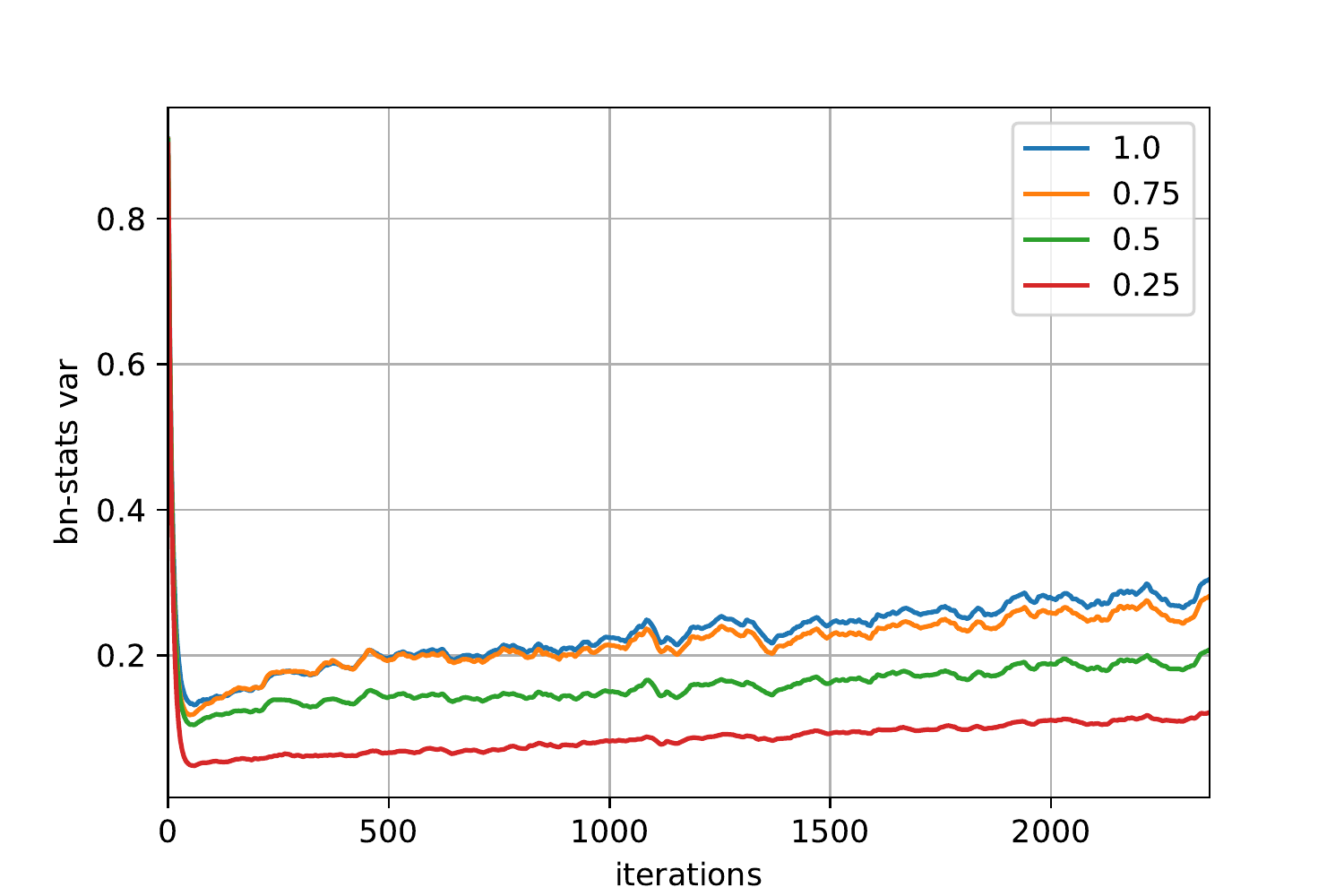}
            \caption{Switchable BN}
            \label{fig:switchbn_stats}
        \end{subfigure}
        ~
        \centering
        \begin{subfigure}[b]{0.32\textwidth}
            \includegraphics[width=\textwidth]{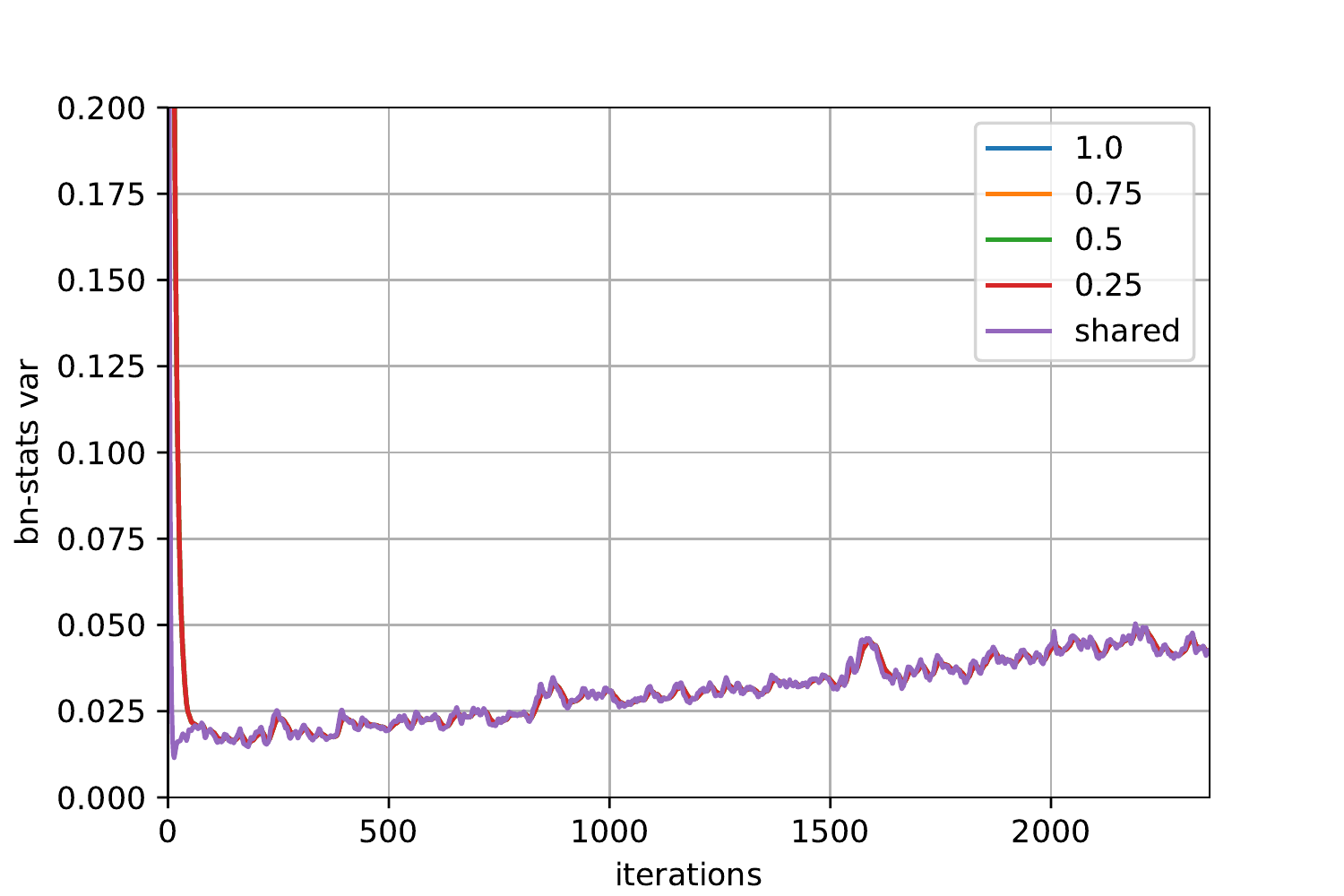}
            \caption{AWN}
            \label{fig:triangular_stats}
        \end{subfigure}
    \end{minipage}
\caption{Tracking the activation means (top row) and variances (bottom row) at 4 width factors of LeNet-3C1L as a function of training iterations. Column (a) uses traditional batch normalization layers, column (b) uses switchable batch normalization \citep{slimmable, slimmable_v2}, and column (c) uses our proposed AWN architecture. The ``shared'' line on the plots in columns (a) and (c) indicate the statistics tracked by a traditional batch normalization layer shared across all widths.
Note how AWN statistics do not deviate from this line at different widths. 
}
\label{fig:varying_stats}
\end{figure*} 

\subsection{Varying statistics in action}\label{section:multi-width-issue}
The variation of batch statistics is empirically visible when training multi-width networks. To demonstrate it, we train and test a variation of the LeNet architecture \cite{lenet} on the MNIST handwritten digits dataset and tracking the activation statistics during training. Our model, LeNet-3C1L, has 3 convolutional layers with 32 neurons each, followed by a single fully connected layer. Every convolutional layer is followed by a traditional batch normalization layer. 

We train LeNet-3C1L with 4 width-factors, $\alpha \in \braces{0.25, 0.5, 0.75, 1.0}$, similar to \citet{slimmable}. All 4 width configurations are trained simultaneously as sub-networks using unweighted sum of losses. To examine the effect of multi-width training on activation statistics, this network uses traditional batch normalization, which shares the same set of accumulated statistics across all widths. During training, we track the means and variances for each feature at each width and compare them to the running statistics being accumulated by the batch normalization layer. Given our analysis in Section \ref{sec:why-stats-vary}, we expect to see a noticeable deviation in the statistics between widths. Sure enough, Figure \ref{fig:sharedbn_stats} shows four distinct sets of activation statistics when switching between the four different width configurations. Even though the distributions all fluctuate over the course of training, the variation between them stays fairly consistent and shows no sign of convergence.

Additionally, we are also interested in whether or not this discrepancy also correlates with poor classification performance, as hypothesized by \citet{slimmable}. To answer this question, we also train a Slimmable version of LeNet-3C1L using switchable batch normalization layers \cite{slimmable}. These layers contain a separate batch normalization module for each width. As switchable batch normalization explicitly accounts for the activation statistics at each width, we consider this to be our baseline. The results of this comparison, given in Table \ref{table:sharedbn_acc}, support the hypothesis that the variation in activation statistics impacts network accuracy. Not only is the overall performance lower than the baseline, but we also see that a middle width configuration, $\alpha = 0.75$, actually performs better than the full-width version ($\alpha = 1.0$).

Note that Figure \ref{fig:switchbn_stats} shows varying statistics even when using switchable batch normalization. This confirms that the different width configurations require tracking different sets of statistics. Because switchable batch normalization explicitly models the statistics at each width, validation accuracy is higher and more consistent. However, while this is effective for $N=4$, this approach does not scale well; the number of batch normalization versions to be measured and stored increases linearly with $N$. Therefore, in the next section, we propose a novel architecture, Any-Width Network (AWN), that combats the width-varying batch statistics issue without requiring a separate model for each width configuration.
\section{Any-Width Networks}
Although existing solutions for multi-width network operation have addressed the problem of varying activation statistics, they fall short of guaranteeing functionality at all widths. The \nwidth design has inherent shortcomings in its ability to make inference-time guarantees about network function at a given width. In particular, it creates a chicken-and-egg problem, where the width-accuracy trade-off curve needs to be known in order to determine the $N$ widths to train, but the trade-off curve cannot be generated without having trained all possible widths. To address this paradox, we propose Any Width Networks (AWNs), which circumvent the problem by enabling operation at all widths after a single training routine without requiring different versions of the model for each width.

\newcommand{\imgsizeTOV}{0.31}
\newcommand{\folderTOV}{widthsH}
\newcommand{\vspaceTOV}{}

\begin{figure}[t]
    \centering
    \begin{subfigure}[b]{\imgsizeTOV\linewidth}
        \centering
        \captionsetup{justification=centering}
        \includegraphics[width=\linewidth]{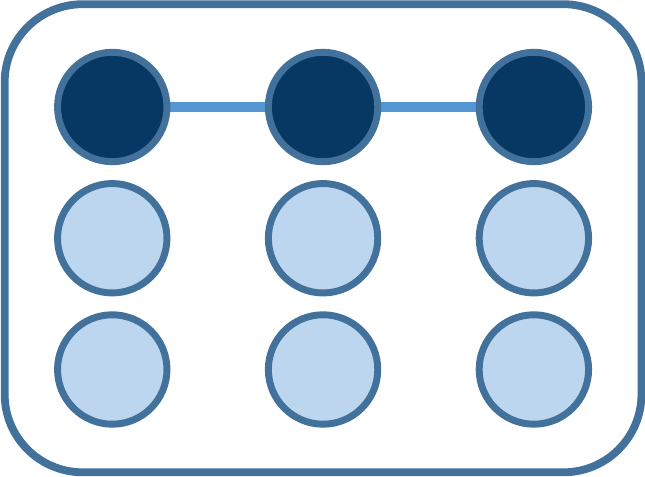}
        \caption*{\textbf{Width = 1}}
    \end{subfigure}
    ~ \vspaceTOV
    \begin{subfigure}[b]{\imgsizeTOV\linewidth}
        \centering
        \captionsetup{justification=centering}
        \includegraphics[width=\linewidth]{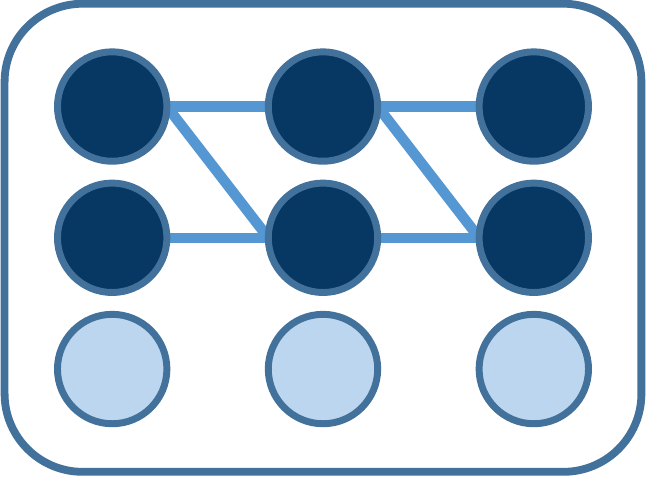}
        \caption*{\textbf{Width = 2}}
    \end{subfigure}
    ~ \vspaceTOV
    \begin{subfigure}[b]{\imgsizeTOV\linewidth}
        \centering
        \includegraphics[width=\linewidth]{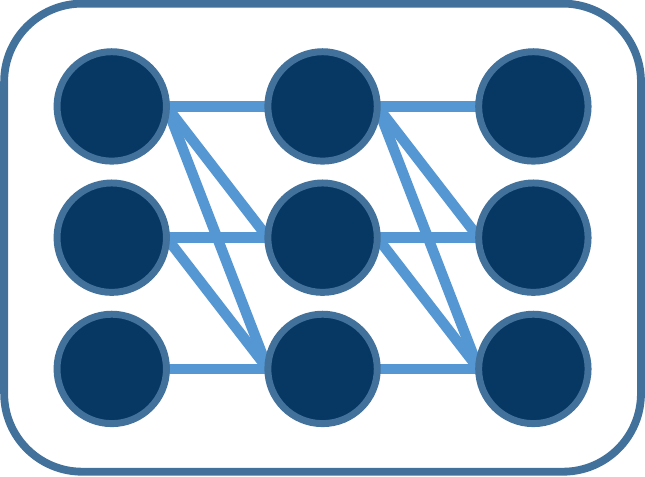}
        \caption*{\textbf{Width = 3}}
    \end{subfigure}
    \caption{AWN operations with different widths.}
    \label{fig:tria-op-vis}
\end{figure}

\subsection{Triangular weight matrices} \label{section:triangular-weight}
To provide multi-width operation without storing separate activation statistics for each width, we must provide an alternative solution to the varying statistics problem. For AWN architecture, our solution is to use lower-triangular weight matrices. Constraining the weight matrices to this form explicitly removes the problem.

Recall our observation in Equation (\ref{eq:expval}) that the only way to ensure that $y_1^{\braces{2}}$ can be modeled by the same distribution as $y_1^{\braces{1}}$ would be to ensure that $\Ebb[w_{12}x_2] = 0$. More generally, we need to be able to guarantee that in a layer operating at width $k_i = k_{i-1}$, for any feature $y_s^{\braces{k}}$, where $s \le k_i$:
\begin{equation}\label{eq:expvaliszero}
    \forall t : t > s \rightarrow \Ebb \left[ w_{st} x_t \right] = 0
\end{equation}
where $w_{st} \in \vect{W}$ and $x_t \in \vect{x}$. Because $\vect{x}$ is a function of the network input and parameters of previous layers, it is impossible to predetermine $\Ebb \left[ w_{st} x_t \right]$ when designing the network. Instead, we achieve this constraint by simply setting $w_{st} = 0$. Doing this for all possible widths produces a lower-triangular weight matrix.
The same idea holds for layers operating at $k_i \neq k_{i-1}$.
A visualization of these triangular matrices at different widths is shown for fully-connected layers in Figure \ref{fig:tria-op-vis}. This design can be used in both fully-connected layers and convolutional layers. A network constructed with these triangular layers can use traditional batch normalization without worrying that multi-width operation will cause deviation from the accumulated statistics, as highlighted in Figure \ref{fig:triangular_stats}. Table \ref{table:sharedbn_acc} shows that when we replace the normal convolution layers with triangular convolution layers in our previous experiment, we also see that our AWN version of LeNet-3C1L has comparable validation accuracy to the version that uses switchable batch normalization \citep{slimmable}.

Note that, in addition to addressing the varying activation statistics problem, the triangular weight matrix design also establishes fully granular control over all widths of the network. As we design the triangular layers around the varying statistics issue, models using these operations do not need to store multiple versions of any layer. As a result, they do not need to be explicitly trained at pre-selected widths and can operate at any desired width. Hence, we refer to a network constructed using triangular layers and traditional batch normalization as an Any Width Network.
\subsection{Random sample training} \label{section:random_training}
\begin{figure}[t]
    \centering
    \includegraphics[width=\linewidth,trim={0cm 0cm 0cm 1.1cm},clip]{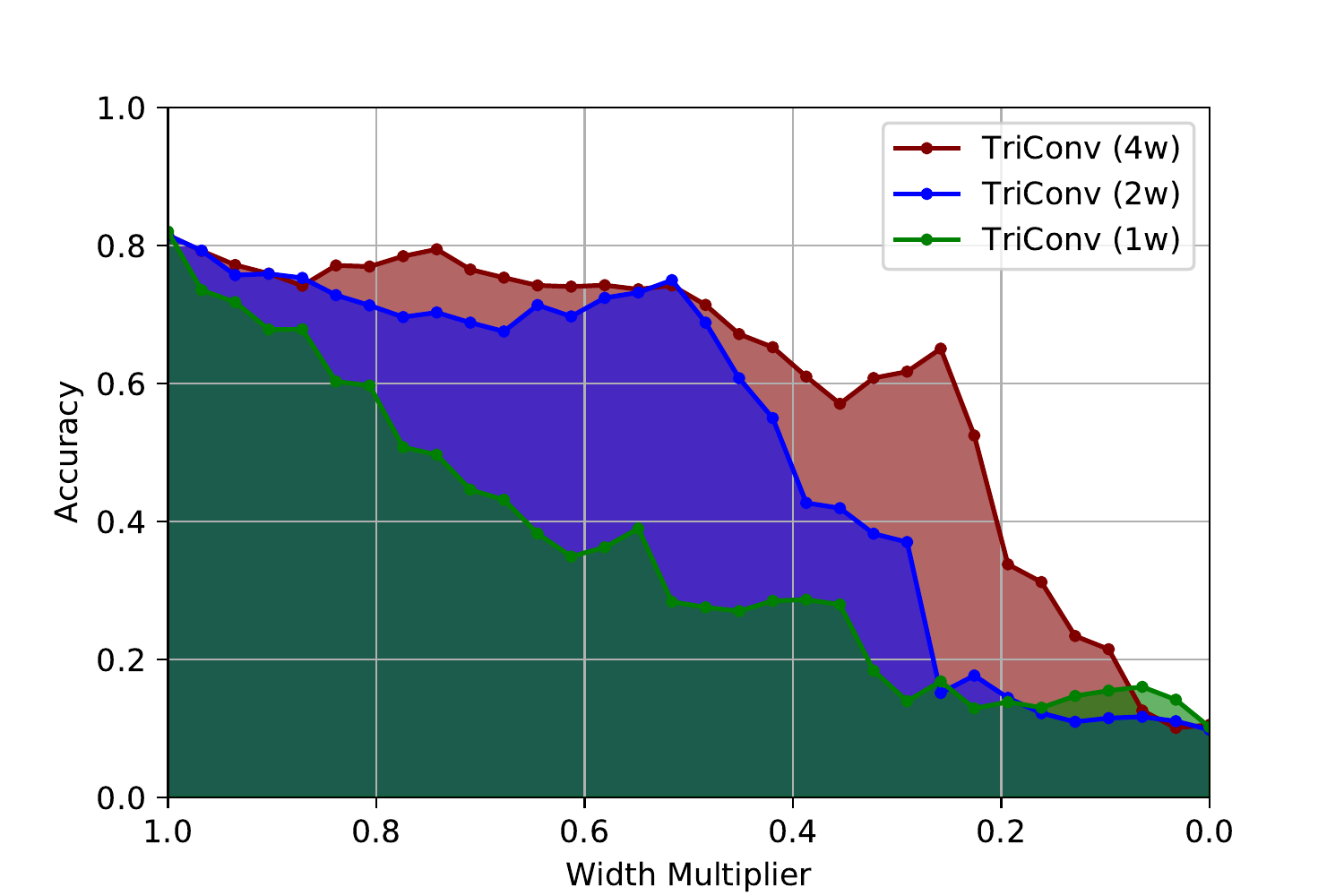}
    \vspace{-6mm}
    \caption{As we add more widths to the training process, we see the area under the width-accuracy trade-off curve increase, indicating better performance. \textit{TriConv (\textbf{n}w)} denotes network trained with \textit{\textbf{n}} widths.}
    \label{fig:random-width-training}
\end{figure}

Intuitively, we expect AWNs to perform best when trained at a variety of widths. To do this efficiently, we employ a random-width sampling strategy to train multiple widths at each training iteration. Rather than perform $N$ explicit forward and backward passes for $N$ widths at each iteration, we randomly sample $n$ width-factors ($n \ll N$) uniformly between a minimum width $\alpha_{\mathrm{min}}$ and maximum width $\alpha_{\mathrm{max}}$. Figure \ref{fig:random-width-training} demonstrates how including more widths during training improves the resulting trade-off curve.

Our approach is similar to that proposed by \citet{slimmable_v2} for Universally-Slimmable Networks. Where our method diverges from theirs, though, is that AWNs do not require any post-training step. Universally-Slimmable Networks require additional processing during training to accumulate activation statistics at each operating width due their use of switchable batch normalization layers. In contrast, AWNs can operate at any width immediately after training.

For the experiments in this paper, we set $n = 4$, and follow the scheme proposed by \citet{slimmable_v2}, where the network is trained using $\alpha_{\mathrm{min}}$, $\alpha_{\mathrm{max}}$, and $(n-2)$ random widths at each iteration. Algorithm \ref{alg:training} outlines our training routine.

\begin{algorithm}[t]
    \caption{$\mathrm{AWN\_RS\_TRAIN}(\mat{A}, N_{\mathrm{iters}}, \alpha_{\mathrm{min}}, \alpha_{\mathrm{max}}, n)$}
    \label{alg:training}
    \begin{algorithmic}[1]
        \REQUIRE $\mat{A}$, any-width network
        \REQUIRE $N_{\mathrm{iters}}$, number of training iterations
        \REQUIRE $\alpha_{\mathrm{min}}$, minimum width-factor
        \REQUIRE $\alpha_{\mathrm{max}}$, maximum width-factor
        \REQUIRE $n$, number of simultaneous widths to train
        \STATE $\mat{A}\mathrm{.initialize}()$
            \FOR{$i = 1, ... N_{\mathrm{iters}}$}
                \STATE Load training data, $\vect{x}$, and labels, $\vect{y^*}$
                \STATE \parbox[t]{\linewidth}{$S =$ sample$\left(\alpha_{\mathrm{min}}, \alpha_{\mathrm{max}}, n \right) \rightarrow$ \\
                $\text{\quad}\{\alpha_{\mathrm{min}}, \alpha_{\mathrm{max}}\}\bigcup\{
                \alpha_i \sim \mathcal{U}\left(\alpha_{\mathrm{min}}, \alpha_{\mathrm{max}}\right)
                \}_{2< i \leq n}$}
                \STATE $\Delta = 0$
                \FOR{$\alpha_j$ in $S$}
                    \STATE $\mat{A}\mathrm{.set\_width}(\alpha_j)$
                    \STATE $\vect{y} = \mat{A}\mathrm{.forward}(\vect{x})$
                    \STATE $L = \mathrm{compute\_loss}(\vect{y}, \vect{y^*})$
                    \STATE $\Delta \mathrel{+}= \mat{A}\mathrm{.backward}(L)$
                \ENDFOR
            \STATE $\mat{A}\mathrm{.grad\_update}(\Delta)$
        \ENDFOR
    \end{algorithmic}
\end{algorithm}
\vspace{-2mm}
\section{Experiments} \label{section:experiments}
We perform two types of experiments. In the first, we look to evaluate the architecture's performance in simulated resource-constrained conditions. For this, we use the image classification task to examine how accuracy is affected as the width of operation is varied. We measure this effect by computing the area under the width-accuracy trade-off curve (AUC) and use this criterion to compare AWNs to prior work. In the second experiment, we aim to understand what makes the AWN network effective for multi-width operation. To do this, we analyze the performance of the triangular convolutional layer at different widths.

\subsection{Width-accuracy trade-off}\label{sec:classification-exp}
\begin{table*}[t] 
\begin{center}
    \begin{tabular}{rr|cc|cccc}
        \hline
        &&\multicolumn{2}{c|}{\textbf{4-Width Training}} &\multicolumn{4}{c}{\textbf{Random Sample Training}} \\
        \textbf{Model} & \textbf{Dataset} & \textit{AWN} & \textit{S-Net} & \textit{AWN+RS} & \textit{US-Net(4w)} & \textit{US-Net(10w)} & \textit{US-Net(20w)}\\
        \hline
        \hline 
        LeNet & F-MNIST & \textbf{90.67}\% & 83.78\% & \textbf{91.06}\% & 90.40\% & 90.70\% & 90.74\% \\
              & CIFAR10  & \textbf{71.12}\% & 56.99\% & \textbf{73.75}\% & 67.84\% & 71.30\% & 72.45\% \\
              & CIFAR100 & \textbf{42.69}\% & 26.94\% & \textbf{43.65}\% & 40.12\% & 41.38\% & 41.80\% \\
        \hline
        MobileNetV2  & CIFAR10 & \textbf{90.91} \% & 50.97\% & 90.72\% & 90.10\% & 91.98\% & \textbf{92.09}\% \\
                     & CIFAR100 & \textbf{70.16}\% & 49.84\% & 67.28\% & 66.69\% & 71.62\% & \textbf{72.24}\% \\
        \hline
    \end{tabular}
\end{center}
\caption{Comparison of AUC performance of our proposed AWNs to Slimmable Networks (S-Net) \citep{slimmable} and Universally-Slimmable Networks (US-Net) \citep{slimmable_v2}. For a fair comparison with the multi-width flexibility of US-Net, we also run their post-training statistics processing routine for 10 and 20 widths, which incur a larger network size cost compared to AWN.
}
\label{table:auc}
\end{table*}
\begin{figure*}[t]
    \centering
        \begin{subfigure}[b]{0.32\textwidth}
            \includegraphics[width=\textwidth,trim={1cm 0 1cm 0},clip]{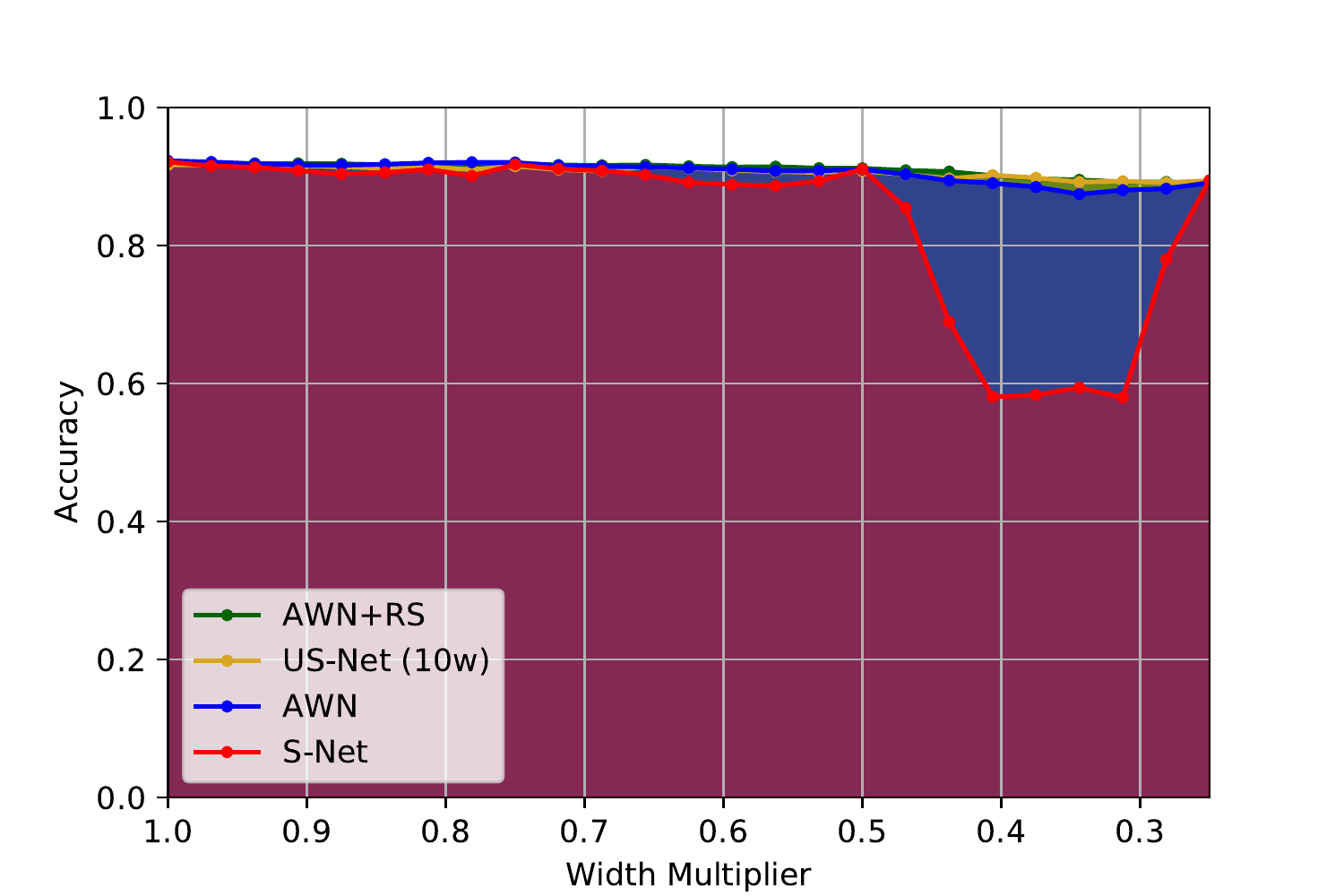}
            \caption*{\textbf{LeNet-3C1L on FashionMNIST}}
        \end{subfigure}
        ~
        \begin{subfigure}[b]{0.32\textwidth}
            \includegraphics[width=\textwidth,trim={1cm 0 1cm 0},clip]{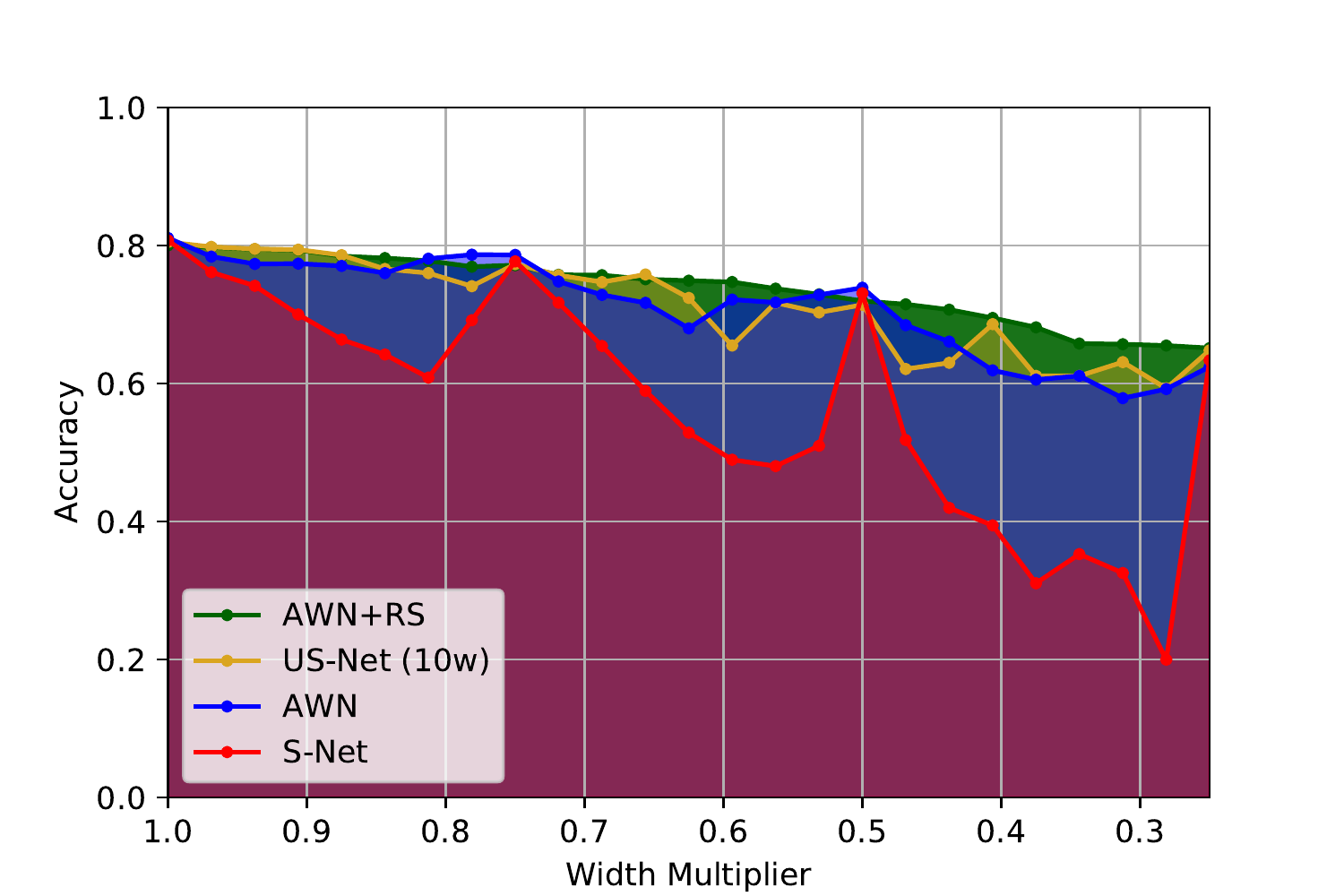}
            \caption*{\textbf{LeNet-3C1L on CIFAR-10}}
        \end{subfigure}
        ~
        \begin{subfigure}[b]{0.32\textwidth}
            \includegraphics[width=\textwidth,trim={1cm 0 1cm 0},clip]{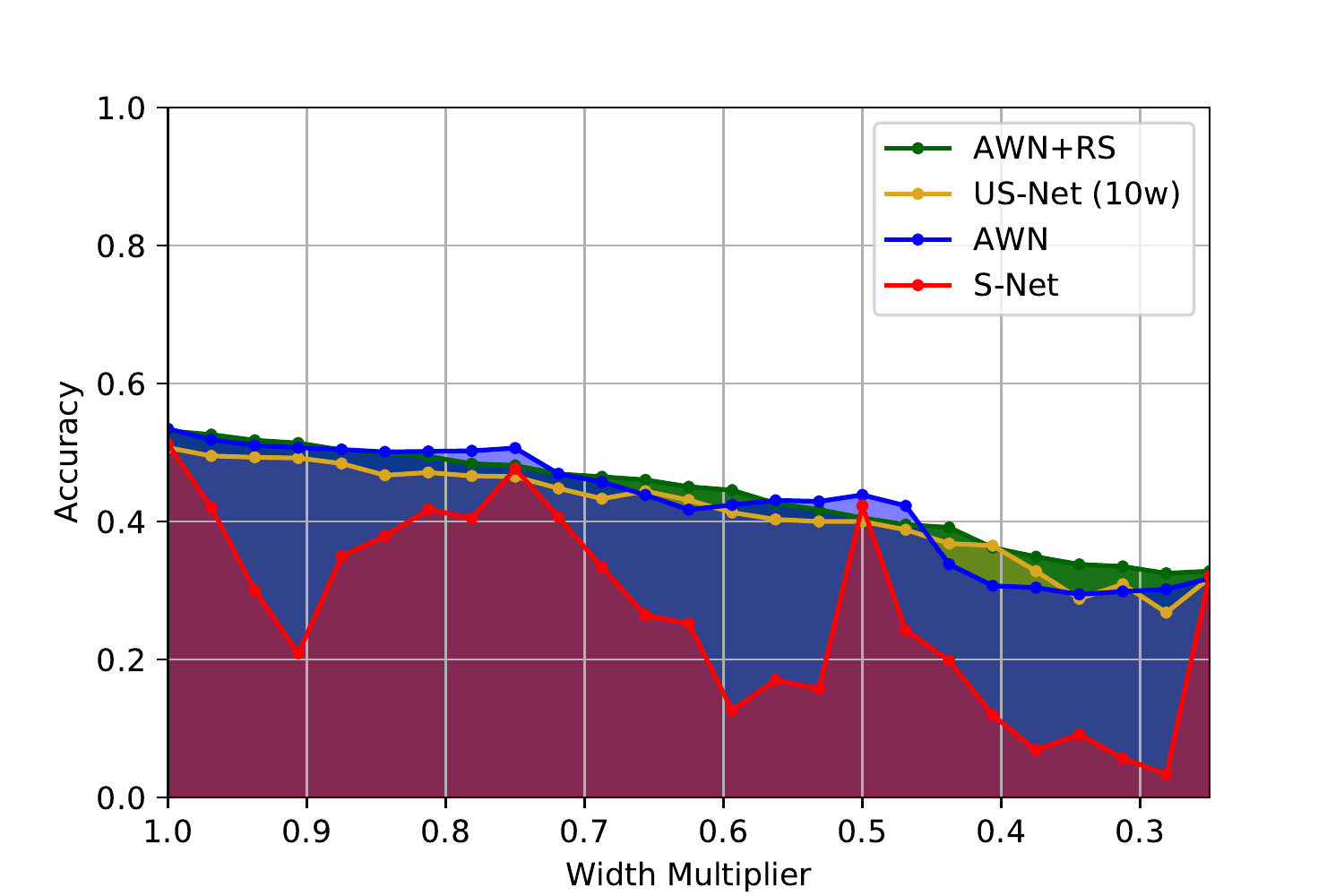}
            \caption*{\textbf{LeNet-3C1L on CIFAR-100}}
        \end{subfigure}\\
        
        
        \begin{subfigure}[b]{0.32\textwidth}
            \includegraphics[width=\textwidth,trim={1cm 0 1cm 0},clip]{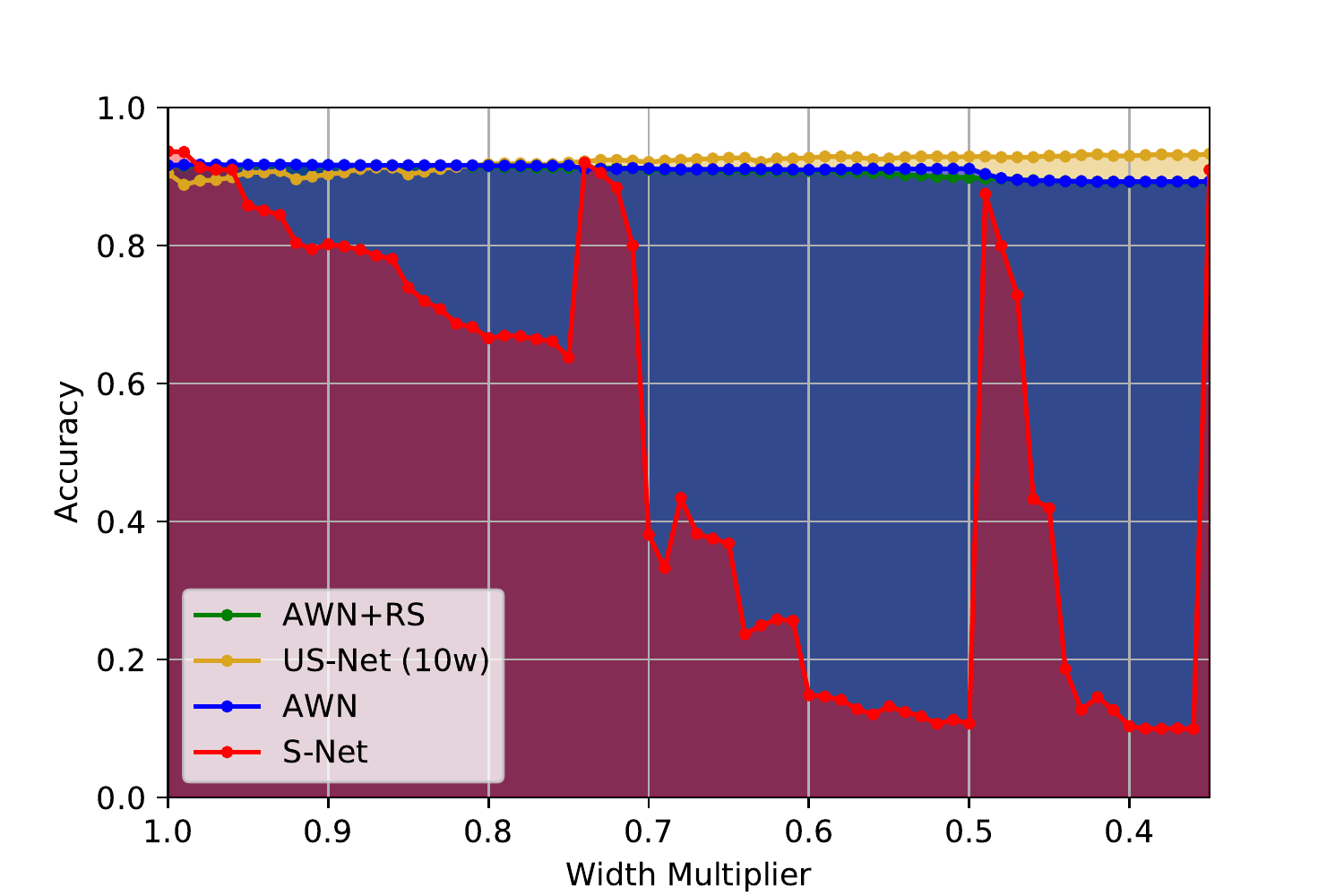}
            \caption*{\textbf{MobileNetV2 on CIFAR-10}}
        \end{subfigure}
        ~
        \begin{subfigure}[b]{0.32\textwidth}
            \includegraphics[width=\textwidth,trim={1cm 0 1cm 0},clip]{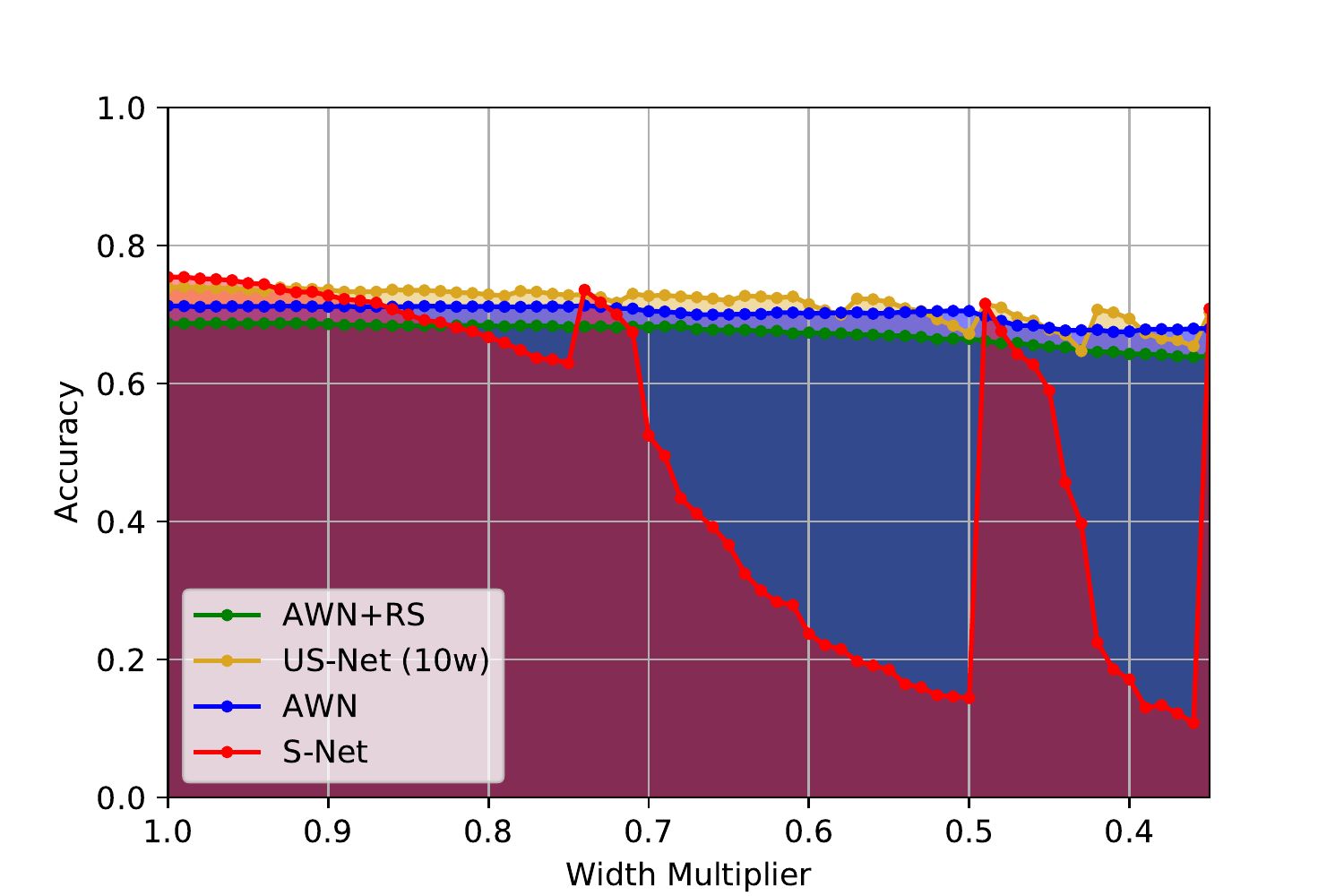}
            \caption*{\textbf{MobileNetV2 on CIFAR-100}}
        \end{subfigure}
        
        \vspace{4mm}
        
\caption{Visualization of the area under the width-accuracy trade-off curves for our classification experiments described in Section \ref{sec:classification-exp}. Our AWN architecture provides a significantly more consistent performance at all widths than S-Net \citep{slimmable}. We achieve similar performance to US-Net \citep{slimmable_v2}, but without the use of switchable batch normalization.}
\label{fig:width-accuracy}
\end{figure*}
We assess performance on three datasets: FashionMNIST \cite{fashion_mnist}, CIFAR-10, and CIFAR-100 \cite{cifar10}. FashionMNIST contains $28 \times 28$ pixel grayscale images labeled according to 10 classes. The dataset is split into training and test sets of 60,000 and 10,000 images, respectively. CIFAR-10 and CIFAR-100 both consist of $32 \times 32$ pixel color images, with 50,000 in each training set and 10,000 in each test set. The datasets differ in that the former has 10 classes, while the latter has 100. On each dataset, we compare our AWN performance to Slimmable Networks (S-Net) \citep{slimmable} and Universally Slimmable Networks (US-Net) \citep{slimmable_v2}, which are both examples of \nwidth architectures.


\textbf{Experimental setup}
For these experiments, we use two base network architectures, LeNet-3C1L and MobileNetV2 \cite{mobilenetv2}, and implement three variants of each: an AWN, an S-Net, and a US-Net.
We widen the AWN variants by a factor of $\sqrt{2}$ so that the number of total network parameters doubles. This is necessary to maintain parity in the number of total active parameters in each model variant, because the triangular weight matrix constraint in AWNs roughly halves the number of parameters used by each layer. For the AWNs, we use traditional batch normalization \cite{batch_normalization} after every convolutional layer. For the S-Net and US-Net variants, we use the relevant switchable batch normalization modules and post-training procedures as described in the original papers. In order to compare resource-constrained performance of the fully-granular AWNs to that of the \nwidth S-Net and US-Net, we must define a selection strategy for $\alpha$ values between the trained widths of the \nwidth models. For this, we select the next-larger sampled width for our desired width of operation. For example, given an S-Net trained at $\alpha_{\text{S-Net}} = \{ 0.25, 0.5 \}$, we would use the batch statistics from $\alpha_{\text{S-Net}} = 0.5$ to evaluate at a width-factor of $\alpha = 0.3$.

All experiments use four width samples per training iteration. For fixed-width training, we set $\{\alpha_i\} = \{1.0, 0.75, 0.5, 0.25\}$ for LeNet-3C1L and $\{\alpha_i\} = \{1.0, 0.75, 0.5, 0.35\}$ for MobileNetV2, following the setup in \cite{slimmable, slimmable_v2}. For random-sample training we follow the sampling strategy described in Section \ref{section:random_training}. We denote the variant of our networks trained with random sampling as AWN+RS. We compare US-Net performance at 4, 10, and 20 post-training width samples, evenly spaced in $[\alpha_{\min},1]$.

\textbf{Training Details}
For reproducibility, we provide all the training parameters used in our experiments. All experiments are trained using SGD using a momentum of $0.9$ and a batch size of $128$. All LeNet-C31L experiments are trained with an initial learning rate of $0.01$, decaying by a factor of $0.1$ at $50\%$ and $75\%$ of the total epochs. The FashionMNIST experiments are trained for $20$ epochs without weight decay, while the CIFAR-10 and CIFAR-100 experiments are trained for $100$ epochs with weight decay $5e^{-4}$. For the MobileNetV2 experiments, our S-Net and US-Net variants are trained for $100$ epochs with a weight decay of $5e^{-4}$ and an initial learning rate of $0.1$, decaying linearly with each iteration. We observe that AWNs do not train as well at such a large learning rate, so we lower it and
increase the training time instead. All AWN experiments, with the exception of CIFAR-100 are trained with a weight decay of $0.001$. The AWN variant is trained on CIFAR-10 for 700 epochs with an initial learning rate of $0.02$, decaying by a factor of $0.2$ at epochs 500 and 600. The AWN+RS variant is trained on CIFAR-10 for 350 epochs with an initial learning rate of $0.01$, decaying by a factor of $0.1$ at epochs 250 and 300. The AWN variant is trained on CIFAR-100 for $100$ epochs with a weight decay of $5e^{-4}$ and an initial learning rate of $0.1$. Finally, the AWN+RS variant is trained on CIFAR-100 for 1050 epochs with initial learning rate of $0.01$, decaying by a factor of $0.1$ at epochs 750 and 900. 

\textbf{Results and analysis}
Results of all experiments are given in Table \ref{table:auc} and Figure \ref{fig:width-accuracy}. In most cases, our AWN performs as well or better than its Slimmable Network and Universally-Slimmable Network counterparts. There are a few points worth discussing further. 

While US-Net performs marginally better than AWN in the MobileNetV2 experiments, this advantage only begins to appear when the number of explicitly trained widths increases. In order to exploit these potential gains, the US-Net must store additional parameters for the switchable batch normalization layers. In the MobileNetV2 experiments, the 10-width and 20-width US-Net models require an additional $9\%$ and $18\%$ more space, respectively, compared to using a traditional batch normalization layer. This limitation of the \nwidth design emphasizes an important advantage of the AWN architecture: AWNs can operate at all possible networks widths, potentially much more than 20, without any impact on the model size.

Furthermore, notice in Figure \ref{fig:width-accuracy} that AWNs produce smooth, consistent width-accuracy trade-off curves, while the S-Nets and US-Nets can have erratic dips, especially at lower widths. This consistent curve is an additional benefit of the AWN design. The accuracy drops are expected behavior for S-Nets and is an acknowledged limitation, according to the authors in their follow-up work \citep{slimmable_v2}. However, US-Nets also suffer from reduced performance at unprocessed widths. This outcome is because, despite the benefit of random width sampling during training, US-Nets are still limited by the $N$ modes of switchable batch normalization during inference. As such, these networks can only hope to achieve a similarly consistent curve by running their post-training statistics accumulation routine for every desired width, with an additional cost of network size.

\subsection{Multi-width suitability}
The Slimmable Network family \citep{slimmable, slimmable_v2} are not inherently suited for the multi-width problem. They are prone to worse performance at widths too far removed from any of their switchable modes. Figure \ref{fig:width-accuracy} illustrates this effect quite clearly: the red curve representing the Slimmable Network drops considerably between the four modes for which it is trained. Although random width-sampling during training and the post-processing of statistics for additional widths of operation, introduced in the Universally-Slimmable Network, appear to be a viable solution for this problem, they do not address the underlying issue that \nwidth design is not naturally ``slimmable.'' Hence, we aim to understand if our AWN design provides any improvement in this regard.

\textbf{Experimental setup}
To evaluate this trait, we compare the performance curve of a LeNet-3C1L network containing triangular convolutional layers, as used in an AWN, to the curve of a standard version using regular convolutional layers. The two networks are trained on CIFAR-10 using the same set of hyperparameters described in Section \ref{sec:classification-exp}, but at full width. We then test both models at different width values to produce the width-accuracy trade-off curves. 

\begin{figure}[t]
    \centering
    \includegraphics[width=\linewidth,trim={0cm 0.1cm 0cm 0.9cm},clip]{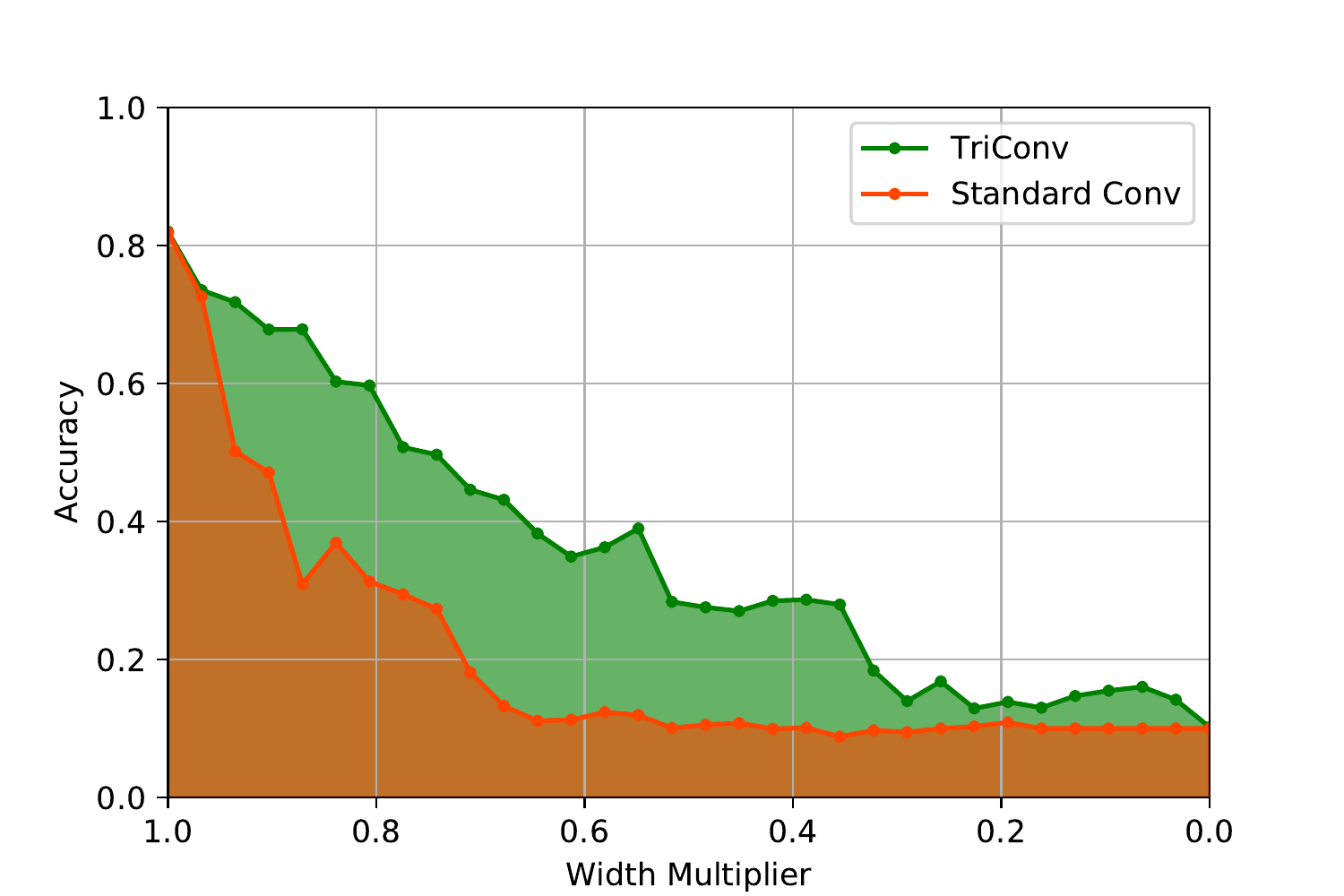}
    \caption{Compared to standard, full-weight matrices (AUC: 20.51\%),  triangular convolutional layers (AUC: 35.85\%) are inherently more suited for multi-width use.}
    \label{fig:naturally-slimmable}
\end{figure}
\textbf{Results and analysis}
The results in Figure \ref{fig:naturally-slimmable} show that our triangular convolutional layer design performs \textbf{75\%} better than standard convolution as the network width is decreased. This graceful performance degradation is more suitable for multi-width applications. We suspect that the source of this benefit is in the triangular design constraint. We hypothesize that constraining each output feature to be a function of a fixed set of input features implicitly imposes a hierarchical importance structure on the features at different widths. This result suggests that triangular convolutional layers may naturally encourage a smooth trade-off curve, even without random width sampling during training. 

\section{Conclusion}
We have proposed Any-Width Networks (AWNs), a family of adjustable-width convolutional neural networks that provide fine-grained control over the width of operation at inference time. We have shown that AWNs sufficiently remove the impact of varying batch statistics on multi-width functionality. We have also demonstrated that AWNs offer maximally granular control with smooth, consistent performance across different widths, without the need to store multiple versions of any network layers. Furthermore, we highlight that the triangular convolutions used in AWNs are naturally suited for multi-width networks. Our results suggest that AWNs provide a promising research direction for resource-constrained inference.
Avenues for future work include applying AWNs to other computer vision tasks such as object detection and segmentation.
\\
\textbf{Acknowledgments} This work was supported by NSF grant CPS 1837337.


{\small
\bibliographystyle{ieee_fullname}
\bibliography{egbib}
}
\end{document}